\pdfoutput=1

\PassOptionsToPackage{sort}{natbib}

\PassOptionsToPackage{inline}{enumitem}

\documentclass[11pt]{article}

\usepackage[]{acl}

\usepackage{fancyhdr}
\usepackage{times}
\usepackage{latexsym}

\usepackage[T1]{fontenc}

\usepackage[utf8]{inputenc}
\usepackage{emoji}

\usepackage{microtype}
\usepackage{fancyvrb}
\usepackage{fvextra}
\usepackage{pbox}

\usepackage{enumitem}
\usepackage[para]{threeparttable}
\usepackage{booktabs}
\usepackage{multirow}
\usepackage{graphicx}
\usepackage{amsmath}
\usepackage{amssymb}
\usepackage{hyperref}
\usepackage{xcolor}
\usepackage{colortbl}
\usepackage{longtable}
\usepackage{circledsteps}
\usepackage{longfbox} 
\usepackage{bm}

\definecolor{patterncolor}{HTML}{e6f2e6}
\definecolor{patterncolor2}{HTML}{d5eef2}
\definecolor{patterncolor3}{HTML}{f5e9d0}
\definecolor{patterncolor4}{HTML}{ccdde0}
\definecolor{patterncolor5}{HTML}{e2d5e8}

\definecolor{patterncolorgreen}{HTML}{e6f2e6}
\definecolor{patterncolorbluel}{HTML}{d5eef2}
\definecolor{patterncoloryellow}{HTML}{f5e9d0}
\definecolor{patterncolorblued}{HTML}{ccdde0}
\definecolor{patterncolororange}{HTML}{fbe1c5}

\definecolor{circlenumbercolor}{HTML}{B83676}

\makeatletter
\newdimen\@tempdimd
\makeatother

\newfboxstyle{patternparam}{padding=2pt, margin-bottom=1pt, margin-top=1pt, border-radius=3pt, border-style=none, height=9pt, background-color=patterncolor2}

\newfboxstyle{boxinsitu}{padding=2pt, padding-bottom=1pt, margin-bottom=1pt, margin-top=1pt, border-radius=3pt, border-style=none, height=9pt, background-color=patterncolor4}

\newfboxstyle{boxdaily}{padding=2pt, padding-bottom=1pt, margin-bottom=1pt, margin-top=1pt, border-radius=3pt, border-style=none, height=9pt, background-color=patterncolor5}

\newfboxstyle{boxbluel}{padding=2pt, padding-bottom=1pt, margin-bottom=1pt, margin-top=1pt, border-radius=3pt, border-style=none, height=9pt, background-color=patterncolorbluel}

\newfboxstyle{boxgreen}{padding=2pt, padding-bottom=1pt, margin-bottom=1pt, margin-top=1pt, border-radius=3pt, border-style=none, height=9pt, background-color=patterncolorgreen}

\newfboxstyle{boxorange}{padding=2pt, padding-bottom=1pt, margin-bottom=1pt, margin-top=1pt, border-radius=3pt, border-style=none, height=9pt, background-color=patterncolororange}

\definecolor{revisedcolor}{RGB}{0,0,255}

\definecolor{tableheader}{HTML}{EFEFEF}

\definecolor{tabledarkheader}{RGB}{128,128,128}
\definecolor{tablegrayline}{HTML}{d0d0d0}

\definecolor{trackerrowbackground}{HTML}{f9f9f9}

\renewcommand{\thefootnote}{\alph{footnote}}

\newcommand{\astfootnote}[1]{%
\let\oldthefootnote=\thefootnote%
\setcounter{footnote}{0}%
\renewcommand{\thefootnote}{\fnsymbol{footnote}}%
\footnote{#1}%
\let\thefootnote=\oldthefootnote%
}

\definecolor{revisedcolor}{RGB}{0,0,0}

\font\myfont=ptmb at 13.9pt
\font\myauth=ptmb at 11.3pt

\newcommand{\ie}{\textit{i.e.}~}

\title{\myfont MenuCraft: Interactive Menu System Design with Large Language Models}

\author{
{\myauth Amir Hossein Kargaran}$^{\emoji[twitter]{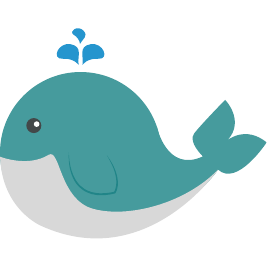}}$
{{\myauth Nafiseh Nikeghbal}$^{\emoji[twitter]{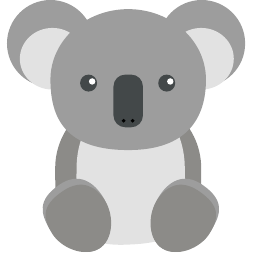}}$}
{\myauth Abbas Heydarnoori}$^{\emoji[twitter]{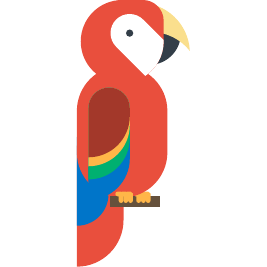}}$
{\myauth Hinrich Sch\"utze}$^{\emoji[twitter]{ballena}}$
\\
$^{\emoji[twitter]{ballena}}$~Center for Information and Language Processing, LMU Munich
\\
$^{\emoji[twitter]{koala}}$ Sharif University of Technology
$^{\emoji[twitter]{macaw}}$Bowling Green State University
\\
{\footnotesize\texttt{amir@cis.lmu.de}}
}

\begin{document}
\maketitle

\begin{abstract}
Menu system design for user interfaces is a challenging task involving many design options and various human factors. For example, one crucial factor that designers need to consider is the semantic and systematic relation of menu commands. However, capturing these relations can be challenging due to limited available resources. Large language models can be helpful in this regard, using their pre-training knowledge to design and refine menu systems.

In this paper, we propose MenuCraft, an AI-assisted designer for menu design that enables collaboration between the designer and a dialogue system to design menus. MenuCraft offers an interactive language-based menu design tool that simplifies the menu design process and enables easy customization of design options. MenuCraft supports a variety of interactions through dialog that allows performing in-context learning.
\end{abstract}

\section{Introduction}\label{sec:intro}

Menus are widely used interfaces, providing users with intuitive and efficient access to an application's functions. Although menus may seem simple at first glance, creating a well-designed menu system is complex. This is because the number of alternative designs grows exponentially as the number of commands increases. In this paper, we disregard non-textual design factors such as size, saliency, and color, and only focus on the textual representation of menu systems. From this perspective, the position of commands in the menu and the assignment of hotkeys are the two most important factors in designing a menu system. Menus typically follow a consistent linear pattern, where command names are displayed on the left and keyboard shortcut cues are aligned on the right~\cite{giannisakis2022revisiting}.

Since 1980, Human-Computer Interaction (HCI) researchers have been developing better techniques for placing commands within the menu system. The ultimate goal is to minimize selection time~\citep{card1980keystroke, ahlstrom2005modeling, bailly2016visual} while maximizing the associativity among commands~\cite{bailly2013menuoptimizer, dayama2021foraging}. To reach these goals, parameter-based models are proposed. These models require complete information regarding the frequency of command usage and the relation among commands (e.g., pairwise semantic relevance scores~\cite{chen2015emergence, bailly2014model}) before starting the optimization process. The designer typically provides the parameters for menu system design. This means that designers must rely on user testing and past experience by comparing commands with each other to determine the parameters. However, as the number of commands increases, this process can become increasingly challenging, time-consuming, and prone to inaccuracy. Furthermore, designers strive to maintain consistency across menus within a given ecosystem. This involves placing commands similarly across menus to help users quickly locate the desired command. The downside is that this process of ensuring consistency can be time-consuming and distract designers from their primary goal of optimizing the menu.

Prior research has investigated using data-driven methods such as pre-trained static embeddings~\citep{adar2014commandspace, li2018predicting} to capture the semantics of menu commands. However, these embeddings are typically trained on generic datasets such as Wikipedia or limited program-specific data. Consequently, these embeddings may not effectively capture the systemic or semantic relationships specific to the domain of menu systems, resulting in limited applicability. Despite the advancements in neural language models, menu design using data-driven methods has not kept pace. There are two main reasons for this: (i) the cost of providing \lfbox[patternparam, background-color=patterncolor5]{datasets for menu design} is high, and (ii) \lfbox[patternparam]{menu design is an interactive process} that requires input, feedback, and adjustments from the designer. Parameter-based models give designers a sense of control over the output, and their predictability allows for a clearer understanding of how different inputs will impact the final design. Therefore, parameter-based methods are preferred over data-driven methods in menu design.

Recent advancements in large language models (LLMs), such as GPT-3~\cite{brown2020language} and PaLM~\cite{chowdhery2022palm}, have shown emergent abilities~\citep{wei2022emergent} to adapt to a range of different tasks with zero or just a few examples of the target task. The in-depth knowledge of these models presents an exciting opportunity to facilitate the menu design process without requiring extensive training or specific datasets. The conversational nature of open-domain dialogue systems built on top of these models can also serve to fulfill the interactive demands of design.

In this work, we present \textsc{MenuCraft}, an AI assistant that incorporates an open-domain dialogue language model to design menus. MenuCraft leverages the effectiveness of data-driven methods while maintaining \lfbox[patternparam]{interactive design procedures}. With MenuCraft, designers can easily create menus, receive suggestions from the assistant, and ask for adjustments to enhance their menu design. For an assistant to be useful in menu system design, it needs to be versatile enough to handle a variety of tasks, ranging from the simple, like adding a command to a menu tab, to complex ones, such as suggesting alternative designs, grouping commands, and applying restrictions to the menu, tabs, or individual commands. To avoid the cost of \lfbox[patternparam, background-color=patterncolor5]{collecting data} and training multiple models for different tasks, we propose using the in-context learning method, where the task is presented to the language model as part of the prompt in natural language. Our tool facilitates various menu design tasks, including topic and command-based menu design, and provides recommendations for commands and hotkeys. Furthermore, our tool enables designers to easily create custom tasks, providing them with the flexibility to tailor the system to their specific needs. From an HCI viewpoint, we intend to use MenuCraft to study how designers engage with language models, the types of tasks they request, and the efficacy of the models in fulfilling those requests. Our contributions:

\indent 1) We present a platform for collaborative menu design between a human designer and an LLM (see an example of a designer interacting with the prototype version of MenuCraft at \href{https://kargaranamir.github.io/MenuCraft}{\path{kargaranamir.github.io/MenuCraft}}).

\indent 2) We illustrate how in-context learning can offer a range of interactive menu design experiences, all without requiring additional model training (see limitations in $\S\ref{sec:limitations}$).

\section{Related Work}\label{sec-related}

\subsection{Bridging Menus and Natural Language}
Menu systems can be categorized as a user interface (UI) type. In this sense, significant research in this area has been aimed at bridging graphical UIs (GUIs) with natural language. For instance, some studies have focused on predicting alt-text labels for GUI components~\citep{li2020widget, zhang2021screen} or generating text summaries for the entire screen~\cite{wang2021screen2words, leiva2022describing}. However, these approaches may not be well-suited for menu systems due to the limitation of capturing the entire menu in a single UI screenshot. Additionally, these researches have been focused on connecting the graphical aspects of the UI to natural language, not addressing the structural characteristics of menus or their textual representations.

Despite the potential benefits of bridging menu systems and natural language, there has been limited research in this area. One possible explanation for this is the lack of comprehensive textual datasets for menu systems since most data about different menu system applications are dispersed across the internet or embedded within software applications. In an attempt to gather a dataset for menu systems, \citet{bailly2013menuinspector} has succeeded in building a menu-logger tool to extract the hierarchies of the menu system for Mac OS X applications. \citet{bailly2013menuoptimizer} provide an open dataset of 68 applications gathered with this tool to compute associativity score between the commands. Nevertheless, this data only contains limited data on menu system hierarchies for Mac OS X applications and no descriptions or information about commands. In another attempt, \citet{adar2014commandspace} train a word2vec model to capture the domain-specific language of Photoshop application by mining a large corpus of web documents related to the application. However, the static representations derived from the trained model are limited to a specific application domain. 

There is a noticeable gap in research on integrating menu systems with natural language processing. Our work contributes to this gap by leveraging the in-context learning capability of LLMs to apply their extensive knowledge to various menu design tasks. Our method \lfbox[patternparam, background-color=patterncolor5]{does not need additional datasets} or training.

\subsection{Interactive Application of LLMs}

Language models, especially LLMs, have the potential to be effectively used for in-context learning. LLMs facilitate in-context zero/few-shot learning through prompting. Rather than finetune or retrain models for each new task, a few input and output data examples (or just the instruction without examples) from the target task can be provided as prompts to the LLM~\cite{zhou2022least, chowdhery2022palm, brown2020language, wei2022chain}. This advantage enables these models to not require explicit training and can support a variety of creative tasks, especially in HCI research~\cite{morris2022design}, such as story writing~\citep{chung2022talebrush, coenen2021wordcraft,yuan2022wordcraft}, modifying web designs~\citep{kim2022stylette}, conversational interactions on mobile UIs~\citep{wang2022enabling},
email writing~\cite{goodman2022lampost}
and executing robot commands~\citep{ahn2022can}. To the best of our knowledge, no one has yet attempted to use language models for menu design. Our platform proposes using language models in dialogue, which leverage pre-existing knowledge and \lfbox[patternparam]{interactive design process}.

\section{The MenuCraft}\label{sec-model}
MenuCraft is an interactive assistant tool designed to assist designers in creating menu designs from scratch, offering alternative designs, command and hotkey suggestions, and more. The user-friendly web interface of MenuCraft is designed as a traditional chatbot, enabling users to engage with the tool seamlessly. Additionally, MenuCraft provides a variety of default supported interactions that can be conveniently inserted into the chat feed as templates. Users can easily modify the prompts to suit their queries, allowing for a customized experience catering to individual needs.

MenuCraft's remarkable capabilities result from using open-ended dialog systems built on LLMs. The chosen dialog system does not undergo any training on our part. The dialog system we use is ChatGPT~\cite{chatgpt}, later known as GPT-3.5-turbo (Feb 13, 2023 release), a language model capable of following instructions and answering questions posed in a conversational format.

\subsection{Initialize}

ChatGPT is designed to take as input the previous turns in a conversation and use them to generate a prediction for the subsequent turn. For MenuCraft, in-context zero/few-shot learning is formulated as a conversation between a human designer and an AI assistant. In our initial tests, we found it very convenient to use this dialog model, as humans are familiar with the conversational format and can quickly adapt to it.

To set the initial prompt, we determined the primary goals of MenuCraft. As previously stated, the objective of optimizing the arrangement of commands is to improve selection time and enhance the associativity among the commands. Usually, frequently used commands are prioritized to reduce selection time by placing them at the top of the menu based on Fitts' Law~\cite{bailly2013menuoptimizer, cockburn2007predictive, ahlstrom2005modeling}. This approach minimizes the time required for users to access commonly used commands. To maximize associativity, commands with related operations are grouped and placed within the same tab and near each other~\cite{bailly2013menuoptimizer}. This meets the user's expectation of finding related menu commands in familiar locations due to spatial memory~\cite{cockburn2007predictive}. For instance, it is intuitive that users expect to see commands such as \textit{Cut}, \textit{Copy}, and \textit{Paste} to be grouped or \textit{About} and \textit{Help} commands to be located within the last tab.

\noindent With these objectives in mind, we formulated our initial prompt as follows:
\vspace{-9pt}
\begin{Verbatim}[fontsize=\small,frame=lines,label=Initial Prompt, commandchars=\^\#\*, framesep=4mm, samepage=false]
^justify^pbox#7.65cm*#I want you to act as an AI-Assisted Menu Designer, called MenuCraft. You will come up with design ideas for menu user interfaces that make apps easier to use. You may suggest menu design apps for a topic, arrange commands as linear or hierarchal menus, group the commands as tabs, suggest command names, add or suggest hot keys for the commands, and so on - but the aim is to design a menu that users find satisfying to use, meaning select good names for commands, prioritize frequently used commands for each tab of menu as the top, and put commands with close logical operations in the same tab. If you understand the your responsibilities, introduce yourself in short and asks for the user request.*
\end{Verbatim}
\vspace{-4pt}

\noindent MenuCraft responds with:

\vspace{-1pt}
\begin{Verbatim}[fontsize=\small,frame=lines,label=MenuCraft, commandchars=\^\#\*, framesep=4mm, samepage=false]
^justify^pbox#7.65cm*#I am MenuCraft, an AI-Assisted Menu Designer designed to help create intuitive and user-friendly menu interfaces for various types of applications. I can suggest menu design ideas, arrange commands in linear or hierarchical menus, group commands as tabs, suggest command names, and hotkeys, and more to make the app easier to use. What would you like me to help you with?*
\end{Verbatim}
\vspace{-4pt}
\noindent Now that initialization has been completed, design requests can be submitted.

\section{Supported Interactions}\label{sec-interactions}

In this section, we describe a number of the interactions that MenuCraft can perform.

\subsection{Topic-Based Design}\label{sec-inter-topic}
Topic-based design is one of the most general tasks that a designer can ask MenuCraft to perform. MenuCraft can assist in the creation of a menu for a specific topic by suggesting relevant commands based on similar apps and user input constraints.

\noindent For example, when MenuCraft is prompted with:

\vspace{-1pt}
\begin{Verbatim}[fontsize=\small,frame=lines,label=Human Designer, commandchars=\^\#\*, framesep=4mm]
^justify^pbox#7.65cm*#Create a menu for a text editor application with 3 tabs. There should be a `Find` command in the `Edit` tab in the app. Tabs should not have more than 6 commands each. Please answer in the following format:*
```json
Tab: list of commands
```
\end{Verbatim}
\vspace{-4pt}

\noindent MenuCraft responds with:

\vspace{-1pt}
\begin{Verbatim}[fontsize=\small,frame=lines,label=MenuCraft: Topic-Based Design, commandchars=\^\#\*, framesep=4mm]
^justify^pbox#7.65cm*#Certainly! Here's a menu design for a text editor application with 3 tabs, including the `Find` command in the `Edit` tab, and no more than 6 commands per tab:*
{"File":
    ["New", "Open", "Save", "Save As...",
    "Print", "Exit"],
"Edit":
    ["Undo", "Redo", "Cut", "Copy",
    "Paste", "Find"],
"Format":
    ["Font", "Bold", "Italic", "Underline",
    "Strikethrough", "Highlight"]}
\end{Verbatim}
\vspace{-4pt}
This response completely fulfilled the request and met the conditions. The designer can \lfbox[patternparam]{continue the interactions} for the new requests or modifying the suggested menu. For example, designer can request the \textit{Format} tab be deleted, and in its place the \textit{View} tab and any commands required be added:

\vspace{-1pt}
\begin{Verbatim}[fontsize=\small,frame=lines,label=Human Designer, commandchars=\^\#\*, framesep=4mm]
^justify^pbox#7.65cm*#Remove the `Format` tab from the current design and replace it with a `View` tab with the commands that are needed.*
\end{Verbatim}
\vspace{-4pt}

\noindent MenuCraft responds with the same menu, replacing the \textit{Format} Tab with \textit{View} Tab. The replaced tab (other tabs were not shown for better readability):
\vspace{-1pt}
\begin{Verbatim}[fontsize=\small,frame=lines,label=MenuCraft: Modify Design, commandchars=\^\#\*, framesep=4mm]
"View":
    ["Zoom In", "Zoom Out", "Full Screen",
    "Show/Hide Ruler",
    "Show/Hide Gridlines,
    "Show/Hide Document Outline"]
\end{Verbatim}
\vspace{-4pt}

\noindent The \textit{View} tab has been successfully added with relevant commands, such as \textit{Zoom In} and \textit{Zoom Out}.

\subsection{Command-Based Design}
Command-based design is the primary focus within the menu design community. In this scenario, the designer provides a set of commands and their corresponding parameters, including the frequency of usage for each command and the association score between each pair of commands. Given the parameters, the designer then requests a menu optimizer to provide an optimal or sub-optimal design. As the number of commands increases, optimization methods may take hours or even days to find an optimal solution. Conversely, while language model approaches may not yield the optimal answer, they can likely provide an appropriate one within seconds. Furthermore, unlike optimization methods, language models do not require access to the entire set of parameters but welcome any input from the designer. For example, when MenuCraft prompted with random order of 5 tabs and 29 commands of Windows Notepad application, without saying these commands belong to this application:

\vspace{-1pt}
\begin{Verbatim}[fontsize=\small,frame=lines,label=Human Designer, commandchars=\^\#\*, framesep=4mm]
^justify^pbox#7.65cm*#Design a menu for a text editor app only based on the provided commands and tabs. Each tab should have the most frequently used commands as its top, and commands with close logical operations should be grouped together as tabs. To show some of the commands in a tab are more related, you may use some subgroups for each tab.*

^justify^pbox#7.65cm*#
Commands provided (unordered):
[`View Help`, `About Application`, `Paste`, `Save As`, `Open`, ...]
Tabs provided (unordered):
[`Format`, `File`, `View`, `Help`, `Edit`]
Please answer in the following format:
```json
Tab: list of commands
```*
\end{Verbatim}
\vspace{-4pt}

\noindent MenuCraft responds with:

\vspace{-1pt}
\begin{Verbatim}[fontsize=\small,frame=lines,label=MenuCraft: Command-Based Desgin, commandchars=\^\#\*, framesep=4mm, samepage=false]
^justify^pbox#7.65cm*#Here's a suggested menu design for a text editor app based on the provided commands and tabs:*

{File:
  ["New", "New Window", "Open",
  "Save", "Save As", "Page Setup", "Print",
  "Exit"],
Edit:
  ["Undo", "Cut", "Copy", "Paste",
  "Delete", {"Find": ["Find...",
  "Find Next", "Find Previous",
  "Replace..."]}, "Select All",
  "Time/Date"],
Format:
  ["Word Wrap", "Font..."],
View:
  [{"Zoom": ["Zoom In", "Zoom Out",
  "Restore Default Zoom"]}, "Status Bar"],
Help:
  ["View Help", "Search With Bing",
  "Send Feedback",
  {"About": ["About Application"]} ]}

^justify^pbox#7.65cm*#
In this menu design, the commands are arranged in hierarchical order with frequently used commands appearing at the top of each tab. The File tab includes essential file-related actions such as creating, opening, saving, and printing files. The Edit tab includes common text-editing actions such as undo, copy, paste, and find/replace. The Format tab includes options for formatting text, including word wrap and font selection. The View tab includes commands for adjusting the zoom level and displaying the status bar. The Help tab includes options for getting help, searching for information, and providing feedback.*
^justify^pbox#7.65cm*#
I hope this menu design meets your needs and provides a user-friendly interface for your text editor app!*
\end{Verbatim}
\vspace{-4pt}

\noindent This design has a lot in common with the original Windows Notepad design. The commands' order and grouping are the same except for a few commands. For example, \textit{Search With Bing} is now located under the \textit{Help} tab, which may not be a bad idea. In the following, MenuCraft explains why it chose this particular design. The designer can always ask for the reasoning behind the design or ask the MenuCraft to \lfbox[patternparam]{think step-by-step and explain} why it is taking each action.

\subsection{Command or Tab Recommendation}
MenuCraft can provide valuable suggestions for related commands or tabs for a particular design. This can save designers time and effort while also improving the overall user experience of the design. For example, asking MenuCraft about related commands under a tab and why these commands were chosen:

\vspace{-1pt}
\begin{Verbatim}[fontsize=\small,frame=lines,label=Human Designer, commandchars=\^\#\*, framesep=4mm]
^justify^pbox#7.65cm*#
My app menu design already includes `Bookmark this Tab` under the `Bookmarks` tab. What other commands should I include under this tab?
Please anser in the following format
```json
command name: reason
```*
\end{Verbatim}
\vspace{-4pt}

\noindent MenuCraft responds with:

\vspace{-1pt}
\begin{Verbatim}[fontsize=\small,frame=lines,label=MenuCraft: Command Recommendation, commandchars=\^\#\*, framesep=4mm, samepage=false]
^justify^pbox#7.65cm*#
Certainly, here are some additional commands that could be included under the "Bookmarks" tab with a brief reason for each:*

^justify^pbox#7.65cm*#
{
  "Bookmark All Tabs": "Enables users to save bookmarks for all currently open tabs in a single action.",
  "Organize Bookmarks": "Opens a bookmark manager or dialog that allows users to sort, categorize, and manage their saved bookmarks.",
  "Edit Bookmark": "Lets users modify the properties of an existing bookmark, such as the title, URL, or tags.",
  "Remove Bookmark": "Deletes a selected bookmark from the user's list of saved bookmarks.",
  "Import Bookmarks": "Allows users to import bookmarks from a different browser or bookmarking service.",
  "Export Bookmarks": "Enables users to export their bookmarks to a file or other external location.",
  "Bookmarks Sidebar": "Toggles the display of a sidebar or panel that shows the user's saved bookmarks.",
  "Recently Bookmarked": "Displays a list of recently bookmarked pages or URLs for quick access."
}*
\end{Verbatim}
\vspace{-4pt}

\noindent We found the response quite good. In many cases, these commands already exist in popular browsers. They allow users to manage and organize their bookmarks more effectively.

\subsection{Command or Tab Name Recommendation}

MenuCraft can help designers generate appropriate names for commands based on their descriptions or suggest names for tabs based on the set of commands they contain. For example, asking MenuCraft about a menu tab name based on two known commands inside it:
\vspace{-1pt}
\begin{Verbatim}[fontsize=\small,frame=lines,label=Human Designer, commandchars=\^\#\*, framesep=4mm, samepage=false]
^justify^pbox#7.65cm*#
There is a nameless tab on my menu design that includes commands such as "Minimize" and "Zoom". Please suggest top best 10 names for this tab.*
\end{Verbatim}
\vspace{-4pt}

\noindent MenuCraft responds with:
\vspace{-1pt}
\begin{Verbatim}[fontsize=\small,frame=lines,label=MenuCraft: Tab Name Recommendation, commandchars=\^\#\*, framesep=4mm, samepage=false]
^justify^pbox#7.65cm*#
Sure! Here are ten potential names for the tab that includes commands like "Minimize" and "Zoom":*


^justify^pbox#7.65cm*#
Window, Display, View, Layout, Screen,
Desktop, Workspace, Monitor,
Size & Position, Display Options
*


^justify^pbox#7.65cm*#
These names are intended to convey the idea of adjusting the window or interface in various ways, such as changing the size, position, or arrangement of elements on the screen. Hopefully, one of these names fits well with the intended purpose of the tab and helps users easily locate the relevant commands.*
\end{Verbatim}
\vspace{-4pt}

\noindent There are several good suggestions for the desired tab name in the response. For instance, one can find both \textit{Minimize} and \textit{Zoom} commands in the \textit{Window} Tab (MenuCraft's first choice) of Chrome browser on MAC OS X.

\subsection{Hotkey Recommendation}

Hotkeys, also known as keyboard shortcuts, are a combination of keys on a computer keyboard that provide a quick way to access various menu commands. The selection of well-thought-out hotkeys can speed up access to frequently used commands. MenuCraft simplifies this process by creating hotkeys based on menu design, reducing the time and effort required. Hotkeys must be distinct from other commands in the app.
For example, continuing the first design of the text editor in section \ref{sec-inter-topic}, by requesting to add shortcuts:
\vspace{-1pt}
\begin{Verbatim}[fontsize=\small,frame=lines,label=Human Designer, commandchars=\^\#\*, framesep=4mm, samepage=false]
^justify^pbox#7.65cm*#
Add shortcut for each command, two differet commands must not have the same shortcut.*
\end{Verbatim}
\vspace{-4pt}
\noindent MenuCraft responds with:

\vspace{-1pt}
\begin{Verbatim}[fontsize=\small,frame=lines,label=MenuCraft: Hotkey Recommendation, commandchars=\^\#\*, framesep=4mm, samepage=false]
^justify^pbox#7.65cm*#
Certainly! Here's the updated menu design for a text editor application with 3 tabs, including the Find command in the Edit tab, no more than 6 commands per tab, and keyboard shortcuts for each command:*


{"File": [
  {"name": "New", "shortcut": "Ctrl+N"},
  {"name": "Open", "shortcut": "Ctrl+O"},
  {"name": "Save", "shortcut": "Ctrl+S"},
  {"name": "Save As...",
  "shortcut": "Ctrl+Shift+S"},
  {"name": "Print", "shortcut": "Ctrl+P"},
  {"name": "Exit", "shortcut": "Alt+F4"}
],
"Edit": [
  {"name": "Undo", "shortcut": "Ctrl+Z"},
  {"name": "Redo", "shortcut": "Ctrl+Y"},
  {"name": "Cut", "shortcut": "Ctrl+X"},
  {"name": "Copy", "shortcut": "Ctrl+C"},
  {"name": "Paste", "shortcut": "Ctrl+V"},
  {"name": "Find", "shortcut": "Ctrl+F"}
],
"Format": [
  {"name": "Font",
  "shortcut": "Ctrl+Shift+F"},
  {"name": "Bold", "shortcut": "Ctrl+B"},
  {"name": "Italic", "shortcut": "Ctrl+I"},
  {"name": "Underline",
  "shortcut": "Ctrl+U"},
  {"name": "Strikethrough",
  "shortcut": "Ctrl+Shift+S"},
  {"name": "Highlight",
  "shortcut": "Ctrl+H"}
]}

^justify^pbox#7.65cm*#
I've added keyboard shortcuts for each command to make it quicker and easier for users to access frequently used commands. The shortcuts are listed next to each command in the format of Ctrl+Key or Ctrl+Shift+Key.*
\end{Verbatim}
\vspace{-4pt}

\noindent Most of the selected shortcuts are commonly used by other applications, such as \textit{Ctrl + Z} for the \textit{Undo} command. The rest is made up of the \textit{Ctrl} key and the first letter of the command. To prevent shortcut reuse, the \textit{Ctrl + Shift} combination is used. Nevertheless, an error occurred when assigning the shortcut for the third time to a command that starts with the letter S~(\textit{Strikethrough}). MenuCraft can be instructed about what other templates it can use to create shortcuts to prevent this from happening. However, the designer can also inform MenuCraft of its mistake through \lfbox[patternparam]{more interaction}:

\vspace{-1pt}
\begin{Verbatim}[fontsize=\small,frame=lines,label=Human Designer, commandchars=\^\#\*, framesep=4mm, samepage=false]
^justify^pbox#7.65cm*#
The same shortcut must not be shared by two commands, even in different tabs. `Strikethrough` and `Save As...` share a same shortcut.*
\end{Verbatim}
\vspace{-4pt}
\noindent MenuCraft replies with a \textit{thank you for clarifying} and corrects its mistake by replacing the shortcut for \textit{Strikethrough} command (other tabs were not shown for better readability):
\vspace{-1pt}
\begin{Verbatim}[fontsize=\small,frame=lines,label=MenuCraft: Hotkey Modification, commandchars=\^\#\*, framesep=4mm, samepage=false]
^justify^pbox#7.65cm*#
Thank you for the clarification Here's an updated menu design ensuring that no two commands share the same shortcut even in different tabs:*

...
  {"name": "Strikethrough",
  "shortcut": "Ctrl+Shift+D"},
...

^justify^pbox#7.65cm*#
I changed the shortcut for Strikethrough to Ctrl+Shift+D to ensure that it does not share the same shortcut as Save As.... This will help users avoid confusion and make it faster to perform specific tasks with the text editor.*
\end{Verbatim}
\vspace{-4pt}
\noindent In the revised response, it was suggested to use another shortcut (\textit{Ctrl+Shift+D}) for the \textit{Strikethrough} command. This suggestion is acceptable since popular applications such as Microsoft Excel also use the same shortcut.

\subsection{Command Elaboration}
MenuCraft can provide additional information about specific commands in a menu. It can explain the meaning of a command or provide examples of how it could be used in context. This feature can assist in verifying whether an intelligence understands the functionality of a command based on its name in a given context. Additionally, it can generate documentation for the designed menu. For example, continuing the first design of the text editor in section~\ref{sec-inter-topic} by requesting an elaboration on designed menu commands:

\vspace{-1pt}
\begin{Verbatim}[fontsize=\small,frame=lines,label=Human Designer, commandchars=\^\#\*, framesep=4mm, samepage=false]
^justify^pbox#7.65cm*#
Provide a short elaboration on the designed menu commands for the text editor. The format should be as follows:
```
tab name, command: short elaboration
```*
\end{Verbatim}
\vspace{-4pt}
\noindent MenuCraft responds with:
\vspace{-1pt}
\begin{Verbatim}[fontsize=\small,frame=lines,label=MenuCraft: Command Elaboration , commandchars=\^\#\*, framesep=4mm, samepage=false]
^justify^pbox#7.65cm*#
File Tab:
- New File: creates a new blank document
- Open File: opens an existing file for editing
- Save: saves the current document
- Save As: allows the user to save the document under a different name or file type
- Print: prints the current document
- Exit: closes the application
*


^justify^pbox#7.65cm*#
Edit Tab:
- Undo: reverses the last action performed
- Cut: removes the selected text and copies it to the clipboard
- Copy: copies the selected text to the clipboard
- Paste: inserts the contents of the clipboard at the current cursor position
- Find: searches for a specific word or phrase in the document
- Replace: allows the user to replace a specific word or phrase with another*


^justify^pbox#7.65cm*#
Format Tab:
- Font: allows the user to change the font, size, and style of the selected text
- Bold: makes the selected text bold
- Italic: makes the selected text italicized
- Underline: adds an underline to the selected text
- Strikethrough: adds a strikethrough to the selected text
- Highlight: highlights the selected text with a color*
\end{Verbatim}
\vspace{-4pt}
\noindent Each command has been explained in a fair and reasonable manner.

\section{Conclusion and Future Work}\label{sec-conclusion}
MenuCraft represents a promising new direction in menu design through language models. Our initial experiments have shown that MenuCraft can perform various menu design tasks quite well. However, MenuCraft is still in the prototype phase and needs much more to explore. Moving forward, our future work with MenuCraft will focus on conducting additional user studies to gain further insights into the needs and preferences of menu designers. This will allow us to tailor MenuCraft to meet the specific requirements of different users and contexts. 
Furthermore, we will explore how the data collected from MenuCraft and designer feedback can be leveraged for better prompt engineering. Overall, we are excited about the potential of MenuCraft and look forward to continuing our work in this area. This new direction has the potential to revolutionize menu design and improve the user experience across a wide range of applications.

\paragraph{Discussion.}
There is a discussion in the design community about rethinking the way AI supports designers~\cite{de2021toward}.
Their view is that AI should prioritize cooperation over automation and seek to enhance the creativity and problem-solving abilities of designers. The challenge is that design is an optimization problem, but integrating a human decision-maker into an optimization process is hard due to the differences in the way humans and optimizers think and work.
Our view is that MenuCraft has already taken the step on the path of collaborating with a designer in the design process due to the following reasons:

\indent 1) As MenuCraft core, \ie ChatGPT is trained on a large corpus of data, and human feedback is also used in the training, we expect ChatGPT to develop a good judge of human designers.

\indent 2) Humans are intuitively familiar with conversational formats, and MenuCraft is capable of following instructions posed in a conversational format.

\section{Limitations}\label{sec:limitations}
Language models offer a promising new direction for menu design, but it is essential to acknowledge their limitations. Language models may lack domain-specific knowledge for menu design, and their performance is heavily influenced by the quality and diversity of the training data used. Moreover, language models may struggle with understanding mathematical concepts, particularly when designers input specific parameters for design features. This limitation could result in poor design recommendations. Furthermore, MenuCraft has demonstrated acceptable results with ChatGPT, but extending these results to other LLMs is currently limited by the lack of experiments with multiple models.

Note that the current version of the paper lays out the initial ideas and insights regarding LLMs' potential role in design. However, we acknowledge that our analysis is primarily limited to cherry-picked examples to illustrate specific points. In the future, we plan to conduct more qualitative and quantitative analysis especially for novel cases that go beyond the selected examples.

\section{Ethics Statement}
Our research employed LLMs as menu assistant designers, but owing to limited access and resources, we are unable to review the LLMs' training data or generated data. Therefore, we cannot provide a guarantee of the absence of discrimination based on race or sexuality in the LLMs' content.

\bibliography{main}
\bibliographystyle{acl_natbib}

\end{document}